\def\year{2022}\relax
\documentclass[letterpaper]{article} 
\usepackage{aaai22}  
\usepackage{times}  
\usepackage{helvet}  
\usepackage{courier}  
\usepackage[hyphens]{url}  
\usepackage{graphicx} 
\usepackage[dvipsnames]{xcolor}
\usepackage{dirtytalk}
\usepackage[normalem]{ulem}
\usepackage{natbib}  
\usepackage{caption} 
\DeclareCaptionStyle{ruled}{labelfont=normalfont,labelsep=colon,strut=off} 
\usepackage{algorithm}
\usepackage{algpseudocode}
\usepackage{subcaption}
\usepackage{float}

\usepackage{newfloat}
\usepackage{listings}
\usepackage{amssymb}
\floatstyle{ruled}
\newfloat{listing}{tb}{lst}{}
\floatname{listing}{Listing}
\title{Estimating Causal Effects in Networks with Cluster-Based Bandits}
\author{
    Ahmed Sayeed Faruk, Jason Sulskis, Elena Zheleva \\
}
\affiliations{
    Department of Computer Science, University of Illinois Chicago\\
    Chicago, IL, USA\\
    \{afaruk2, jsulsk1, ezheleva\}@uic.edu

}

\usepackage{bibentry}

\begin{document}
\maketitle

\begin{abstract}
The gold standard for estimating causal effects is randomized controlled trial (RCT) or A/B testing where a random group of individuals from a population of interest are given treatment and the outcome is compared to a random group of individuals from the same population. However, A/B testing is challenging in the presence of interference, commonly occurring in social networks, where individuals can impact each others outcome. Moreover, A/B testing can incur a high performance loss when one of the treatment arms has a poor performance and the test continues to treat individuals with it. Therefore, it is important to design a strategy that can adapt over time and efficiently learn the total treatment effect in the network. We introduce two cluster-based multi-armed bandit (MAB) algorithms to gradually estimate the total treatment effect in a network while maximizing the expected reward by making a tradeoff between exploration and exploitation. We compare the performance of our MAB algorithms with a vanilla MAB algorithm that ignores clusters and the corresponding RCT methods on semi-synthetic data with simulated interference. The vanilla MAB algorithm shows higher reward-action ratio at the cost of higher treatment effect error due to undesired spillover. The cluster-based MAB algorithms show higher reward-action ratio compared to their corresponding RCT methods without sacrificing much accuracy in treatment effect estimation.
\end{abstract}

\section{Introduction}
Randomized controlled trials, also known as A/B testing, are widely considered the ”gold standard” in causal inference for measuring the effect of changes. They offer a systematic framework to discover the best treatment effect implementation through estimation of the average treatment effect. Many different treatments are tested through A/B experiments, where one class of users are placed into a treatment group and another class of users are placed into a control group. Technology companies like Google and Facebook run thousands of A/B tests every year in order to make data-driven decisions to improve how they deliver services to their customer base. 

A typical A/B experiment relies on the Stable Unit Treatment Value Assumption (SUTVA), or that the response of an individual sample depends only on its own assignment to a particular treatment group. SUTVA states that every individual’s outcome is affected only by their own treatment and not by the treatment of any other individual \cite{G15}. This assumption, however, falls apart when individuals interact with each other, leading to spillover effects. The problem of spillover the treatment from a treated node to a control node through a shared network edge (e.g., information flowing from person to person in an online social network, and diseases spreading between people interacting in the same physical space) leads to interference, which violates SUTVA and thus results in inaccurate treatment effect estimation \cite{halloran2016}. 

An important estimand of interest in networks is the total treatment effect (TTE) which aims to estimate the difference in outcomes in two alternative worlds, one in which everyone is treated and one in which everyone is not treated. For example, we may be interested in comparing a world in which everyone is vaccinated and a world in which no one is vaccinated. Or we may be interested to compare a world in which everyone sees results from the same news feed algorithm versus a world in which everyone sees an alternative news feed algorithm. In both of these worlds interference is allowed. However, due to the fundamental problem of causal inference, estimating this difference directly is impossible. A prominent RCT design for estimating TTE is cluster-based randomization which separated a network into clusters that densely connected within and sparsely connected across \cite{ugander2013, Fatemi20}. Then it assigns treatment and control randomly based on the clusters, keeping interference within clusters but minimizing potential interference across clusters. However, estimating TTE via cluster-based A/B testing can incur a high performance loss when one of the treatment arms has a poor performance and the test continues to treat individuals with it.

Reinforcement learning (RL), on the other hand, is a class of methods that learn from sequences of actions to optimize long-term reward in a Markov Decision Process through exploration of states given specific actions. The Multi-armed Bandit (MAB) problem is a classic problem in this paradigm, where a limited set of resources must be allocated between alternative choices in a way that maximizes overall expected rewards. In contrast to A/B testing, high-performing arms are rewarded with more traffic, whereas under-performing arms are punished with less traffic throughout the MAB experiment. The main advantage of MAB experiment over A/B testing is that it requires much smaller sample and thus terminates earlier, but at the cost of a larger false positive rate due to the uneven distribution of samples. MAB can be a good fit with proper exploration-exploitation trade-off in very large networks, where A/B test is expensive in estimating causal effect of interest. 

Here, we study the use of multi-armed bandits in cluster-based randomization as a more robust and optimal way for learning the total treatment effect in networks. Our target is to estimate the TTE error and reward-action ratio using two-arm bandit network experiments. We propose two cluster-based and one node-based bandit algorithms for network experiments. All these MAB algorithms aim to assign the best performing arm to the newly arrived nodes, but the cluster-based and CMatch-based MAB algorithms also consider the clusters in the network when assigning an arm. The goal of our work is to evaluate and compare the performance of our MAB experiment designs with the corresponding A/B experiments. We also aim to show that exploring a subset of the original network may suffice to learn the total treatment effect with a better reward-action ratio.


\section{Related Work}


Lattimore et. al. have shown that MAB algorithms can be directly applicable to online learning scenarios that involve causal inference, such as exploring treatment effects \cite{lattimore2016causal}. A noble non-parametric MAB based exploration algorithm for partially observed incomplete networks has been introduced by Madhawa and Murata \cite{Madhawa2019}. 

Sliding window UCB algorithm has been proposed by Garivier and Moulines which considers the most recent observation to address the issue of non-stationarity in the environment \cite{Garivier11}. Bootstrap Thompson sampling method can be computationally efficient in large scale bandit problems to optimize exploration-exploitation trade-off in RL \cite{eckles2014thompson}. Lihong et. al. formulates a feature-based exploration/exploitation problem as a contextual bandit problem where a learning algorithm sequentially selects articles to serve users based on contextual information about the users and articles and adapting its article-selection strategy based on the feedback to maximize total rewards \cite{Li10}. 

In case of uncertainty, the upper confidence bound (UCB) algorithm follows the principle of optimism which implies that we should optimistically assume an action as correct when we are uncertain about that action. The uncertainty in the estimates is represented by the confidence interval. The actual value of an action is near to the estimated value when the confidence interval is small. In case of large confidence interval, we become uncertain that the actual value of action is near to the estimated value. A distributed variant of the well-known UCB algorithm has been developed in a multiplayer network where the players can interact among themselves to maximize the overall network rewards together \cite{Dubey2020}. Auer incorporates an exploration term in multi-armed bandits by calculating the confidence bound for each arm and choosing the action corresponding to the largest confidence bound \cite{Auer02}. We have customized the UCB1 algorithm \cite{audibert2011} and adapted it to our two-arm bandit experiment on different network datasets with simulated interference.

The traditional assumption in causal inference research is no interference, but being motivated by the study of infectious diseases, research in statistics has focused on relaxing this assumption \cite{HALLORAN95,TCHETGEN12,HALLORAN12}. The cluster-based randomization (CBR) experiment is a common approach to minimize the effect of interference where nodes are clustered based on their connections and treatment is assigned at a cluster level \cite{ugander2013}. 

Estimating TTE using cluster-based network experiment reduces interference by partitioning the network into clusters with dense inter-cluster and few intra-cluster edge connections \cite{Fatemi20, ugander2013, Saveski17}. The two-stage randomization is randomizing treatment and control group assignment at the cluster-level and dictating the cluster assignment as the node assignment within each cluster \cite{basse2017analyzing}. CMatch framework incorporates node matching and heterogenic edge spillover probability into graph clustering method to match treatment and control cluster pairs which minimizes both inference and selection bias together in network experiment design \cite{Fatemi20}.
 Ban and He proposes a contextual multi-armed bandit algorithm, LOCB to cluster users based on some unknown bandit parameters, which are estimated in an incremental way\cite{ban2021}. A cluster-based bandit algorithm has been developed to efficiently learn the correct cluster for the new users in the design of a recommender system \cite{shams2021}.

\section{Problem Description}
We define a graph $G = (V, E)$,  consisting of a set of $n$ nodes $V$ and a set of edges $E = \{e_{v_{i}v_{j}}\}$ where $e_{v_{i}v_{j}}$ denotes that there is an edge between node $v_i \in V$ and node $v_j \in V$. The \say{edge spillover probability} $e_{v_{i}v_{j}}.p$, is defined the same way as \cite{inproceedings} and refers to the probability of interference occurring between two distinct nodes.   

We let $v_i.X$ denote the pre-treatment node feature for $v_i$, $v_i.Y$ denote the outcome variable of interest for each node $v_i$, and $v_i.T \in \{0, 1\}$ denote whether node $v_i$ has been treated or not. We assume that both $v_i.T$ and $v_i.Y$ are discrete binary variables, $v_i.T = 1$ indicates treated and $v_i.T = 0$ indicates not treated. $v_i.Y = 1$ indicates activated and $v_i.Y = 0$ indicates not activated. 
\subsection{Estimand of interest}
\subsubsection{Total Treatment Effect}
Let $Z \in \{0, 1\}^n$ be the treatment assignment vector of all nodes. The TTE is defined as the outcome difference between two alternative treatment, one in which all nodes are assigned to treatment $(\mathbf{Z}_1 = \{1\}^{N_1})$ and one in which all nodes are assigned to control $(\mathbf{Z}_0 = \{0\}^{N_0})$. The $TTE$ accounts for two types of effects- individual effects, and spillover effects and is defined as follows:
\begin{equation}
   TTE = \frac{1}{N_1}\sum_{v_i \in V}(v_i.Y(\mathbf{Z_1})) - \frac{1}{N_0}\sum_{v_i \in V}(v_i.Y(\mathbf{Z_0})) 
\end{equation}

\subsection{Preliminaries}
Next, we describe the designs for network A/B test experiments.
\subsubsection{Node-based A/B test design:}
A node-based randomized design takes the whole network, G as the input and assigns the nodes to treatment and control groups randomly. This treatment assignment leaves significant number of edges between the treatment and control groups which results in undesirable spillover and thus inaccurate causal effect estimation.

\subsubsection{Cluster-based A/B test design:}
A cluster-based randomized design takes the whole network, G as an input and finds clusters using a graph clustering algorithm. It assigns the clusters to treatment and control groups randomly. The purpose of graph clustering is to find clusters with high intra-cluster and low inter-cluster edge density \cite{Zhou09}. 

\subsubsection{CMatch based A/B test design:}
Fatemi and Zheleva \cite{Fatemi20} show that cluster-based randomization does not ensure sufficient node randomization and it can lead to selection bias in which treatment and control nodes represent different populations of users. Moreover, clustering without considering “edge spillover probability” may result in the assignment of node pairs with high probability of interaction into different clusters, which may lead to undesired spillover. A CMatch-based randomized design takes the whole network as an input and clusters the graph. After node matching, graph clustering, and cluster matching via CMatch framework \cite{Fatemi20}, we get matched cluster pairs. 
The clusters are matched based on the maximal similarity of their corresponding nodes. Therefore, matching cluster pairs tend to represent similar population of users which help to reduce selection bias in randomization. 


If cluster $c_i$ and $c_j$ are matched, then we call $match[c_i] = c_j$ and $match[c_j] = c_i$. This two-stage randomization framework for network experiment design minimizes both selection bias and interference \cite{Fatemi20}. In this randomized algorithm design, one cluster is assigned to the treatment group and the corresponding matching cluster is assigned to the control group randomly from each matching cluster pair. In both cluster-based and CMatch-based algorithm variants, all the nodes in the same cluster get the same treatment. 

\subsection{Interference}
In TTE estimation for real-world networks, there are two types of interference, allowable and unallowable interference. Allowable interference happens within the same treatment group which is a natural consequence of network interactions. Unallowable interference happens across treatment groups which leads to inaccurate causal effect estimation. Pairwise interference can occur when the treatment of one unit impacts the outcome of another unit (direct interference) or when the outcome of one unit impacts the outcome of another unit (contagion) \cite{ogburn2014}.

There are three ways in which a node can get activated- direct treatment, allowable spillover, and unallowable spillover. When a new node arrives (e.g., a user checking their news feed), it can be activated due to their treatment or interference. If the treated node becomes active, it can activate other nodes as well.
\subsection{Reinforcement learning setup}
Our reinforcement learning algorithm learns the parameter for the two arms, treatment and control with the arrival of nodes. The payoffs or rewards of an arrival node refer to the total number of newly activated nodes due to the chosen arm's action on the arrival node, contagion from the arrival node to the already explored network, and contagion from the already explored network to the arrival node. When a node is activated, a reward of $1$ is incurred; otherwise, the reward is $0$. 

The goal is to learn TTE fast while keeping the error minimum and reward-action ratio maximum. The action is defined as the assignment of a node to a treatment class. This reward-action ratio (R/A) is defined as the ratio of total number of activated node to the total number of explored nodes in the network which can be formulated as follows:
\begin{equation}
   R/A = \frac{1}{N_0 +N_1}\sum_{v_i \in V}(v_i.Y(\mathbf{Z_1}) + v_i.Y(\mathbf{Z_0}))
\end{equation}
where $N_0$ and $N_1$ refer to the total number of nodes assigned to the control arm and treatment arm, respectively. 

Aside from node-based bandit, we are also interested in cluster-based bandit algorithms which is described in the next section.


\begin{algorithm}[tb]
	\caption{Node-based MAB} 
	\begin{algorithmic}[1]
	    \State Inputs: $\alpha \in \mathbb{R}_{+}$, $\mathcal{A}_t$
		\For {nodes arriving at time $t=1,2,3,\ldots,T$}
		    \For {all $a \in \mathcal{A}_t$}
		        \If {$a$ is new}
            		\State {$\hat\mu_a \leftarrow 0$} 
            		\State {$m_a \leftarrow 1$}
        		\EndIf	
				\State $p_{t,a} \leftarrow \hat\mu_a + \alpha \sqrt{\frac{2\ln{t}}{m_a}}$
			\EndFor
			\State Choose arm $a_t = \arg \max_{a \in A_t} p_{t,a}$ with ties broken arbitrarily
			\State Observe the payoff $r_t$ due to the treatment and interference 
			\State $m_{a_t} \leftarrow m_{a_t} + 1$
			\State $\hat\mu_{a_t} \leftarrow \frac{r_t 
			+ (m_{a_t} - 1)\hat\mu_{a_t}}{m_{a_t}}$
			
			\State {Update the metrics of interest}
		\EndFor
	\end{algorithmic} 
\end{algorithm}
\section{Cluster-based bandits}
The A/B test algorithm variants decide before the test starts which parts of the network will be assigned to treatment and which parts to control, and assess the reward-action ratio and treatment effect at the end of the experiment. In contrast, the MAB algorithm decides when a node arrives whether to assign it to treatment or control group. The reward-action ratio and treatment effect error is gradually estimated and updated with the arrival of new nodes. 

In this section, we describe the vanilla node-based MAB algorithm, which is followed by cluster-based and CMatch-based MAB algorithm. 

\begin{algorithm}[tb]
	\caption{Cluster-based MAB} 
	\begin{algorithmic}[1]
	    \State Inputs: $\alpha \in \mathbb{R}_{+}$, $\mathcal{A}_t$
		\For {nodes arriving at time $t=1,2,3,\ldots,T$}
            \State Find the cluster $c_t$ of the node arrived at time $t$
            \If {any node $\in c_t$ is already assigned an arm $a_c$}
                \State $a_t \leftarrow a_c$  
         	\Else
    			\For {all $a \in \mathcal{A}_t$}
    		        \If {$a$ is new}
                		\State {$\hat\mu_a \leftarrow 0$} 
                		\State {$m_a \leftarrow 1$}
        		    \EndIf		
    				\State $p_{t,a} \leftarrow \hat\mu_a + \alpha \sqrt{\frac{2\ln{t}}{m_a}}$
    			\EndFor
    			\State Choose arm $a_t = \arg \max_{a \in A_t} p_{t,a}$ with ties broken arbitrarily
    		\EndIf
    		\State Observe payoff $r_t$ due to the treatment and interference
			\State $m_{a_t} \leftarrow m_{a_t} + 1$
			\State $\hat\mu_{a_t} \leftarrow \frac{r_t 
			+ (m_{a_t} - 1)\hat\mu_{a_t}}{m_{a_t}}$
			\State {Update the metrics of interest}
		\EndFor
	\end{algorithmic} 
\end{algorithm}

\begin{algorithm}[tb]
	\caption{CMatch-based MAB} 
	\begin{algorithmic}[1]
	    \State Inputs: $\alpha \in \mathbb{R}_{+}$, $match$ from CMatch, $\mathcal{A}_t$
		\For {nodes arriving at time $t=1,2,3,\ldots,T$}
            \State Find the cluster $c_t$ of the node arrived at time $t$
            \If {any node $\in c_t$ is already assigned an arm $a_c$}
                \State $a_t \leftarrow a_c$  
         	\Else
         	    \If {any node $\in match[c_t]$ is assigned to any arm}
            		\State $a_t \leftarrow$ complimentary arm
        		\Else
        			\For {all $a \in \mathcal{A}_t$}
        		        \If {$a$ is new}
                    		\State {$\hat\mu_a \leftarrow 0$} 
            		        \State {$m_a \leftarrow 1$}
                		\EndIf	
        				\State $p_{t,a} \leftarrow \hat\mu_a + \alpha \sqrt{\frac{2\ln{t}}{m_a}}$
        			\EndFor
        			\State Choose arm $a_t = \arg \max_{a \in A_t} p_{t,a}$ with ties broken arbitrarily
    			\EndIf
    		\EndIf
    		\State Observe payoff $r_t$ due to the treatment and interference
			\State $m_{a_t} \leftarrow m_{a_t} + 1$
			\State $\hat\mu_{a_t} \leftarrow \frac{r_t 
			+ (m_{a_t} - 1)\hat\mu_{a_t}}{m_{a_t}}$
			\State {Update the metrics of interest}
		\EndFor
	\end{algorithmic} 
\end{algorithm}

\subsection{Node-based MAB}
Our vanilla node-based MAB algorithm is a simple application of the UCB1 algorithm \cite{audibert2011} which we describe next as shown in the Algorithm $1$.

New nodes arrive in discrete time intervals $t = 1, 2, 3,\ldots.$ The $\mu_a$ denotes the estimate of pay-off	due to the assignment of the new node to an arm $a$ and $m_a$ denotes to the total number of nodes assigned to the arm $a$. Both of these parameters are initiated in the beginning of the algorithm [Line: $3-7$]. In time t, the algorithm observes the current node $u_t$ and a set $\mathcal{A}_t$ of arms or actions together for $a \in \mathcal{A}_t$, where $\mathcal{A}_t \in \{treatment, control\}$ (Line: $8$). Based on observed payoffs in previous nodes, the algorithm chooses an arm $a_t \in \mathcal{A}_t$, and receives payoff $r_t$ [Line: $10-11$]. The algorithm then improves its arm-selection strategy with the new observation, $(a_t, r_{t,a_t})$ [Line: $12-13$].

The optimal arm for the arrival node at each trial t is chosen by the following equation:
\begin{equation}
a_t = \arg \max_{a \in \mathcal{A}_t}(\hat\mu_a + \alpha \sqrt{\frac{2\ln{t}}{m_a}})
\end{equation}

The UCB1 algorithm \cite{audibert2011} always chooses the arm with the highest confidence interval. The confidence interval increases with the increase in the total number of arrived nodes, t and shrinks with the increase in $m_a$ and thus makes a balance between exploration and	exploitation.



\subsection{Cluster-based MAB}
Here we describe our cluster-based MAB algorithm as shown in the Algorithm $2$. The graph is clustered first and treatment assignment happens at the cluster level.

When a new nodes arrives, its cluster is checked to find whether any other node in the same cluster is already assigned to an arm [Line: $3$]. If any node in that cluster is already assigned an arm, the newly arrived node is assigned to the same arm [Line: $4-5$]. If no node in that cluster is assigned to any arm, the newly arrived node is assigned to an arm based on Equation -$3$ [Line: $12-14$] and the learning parameters and metrics of interest are updated for the newly arrived node as shown in the Algorithm $2$ [Line: $17-19$]. 
\begin{figure*}[!ht]
     \centering
     \begin{subfigure}{0.33\textwidth}
         \centering
         \includegraphics[width = 2.2 in]{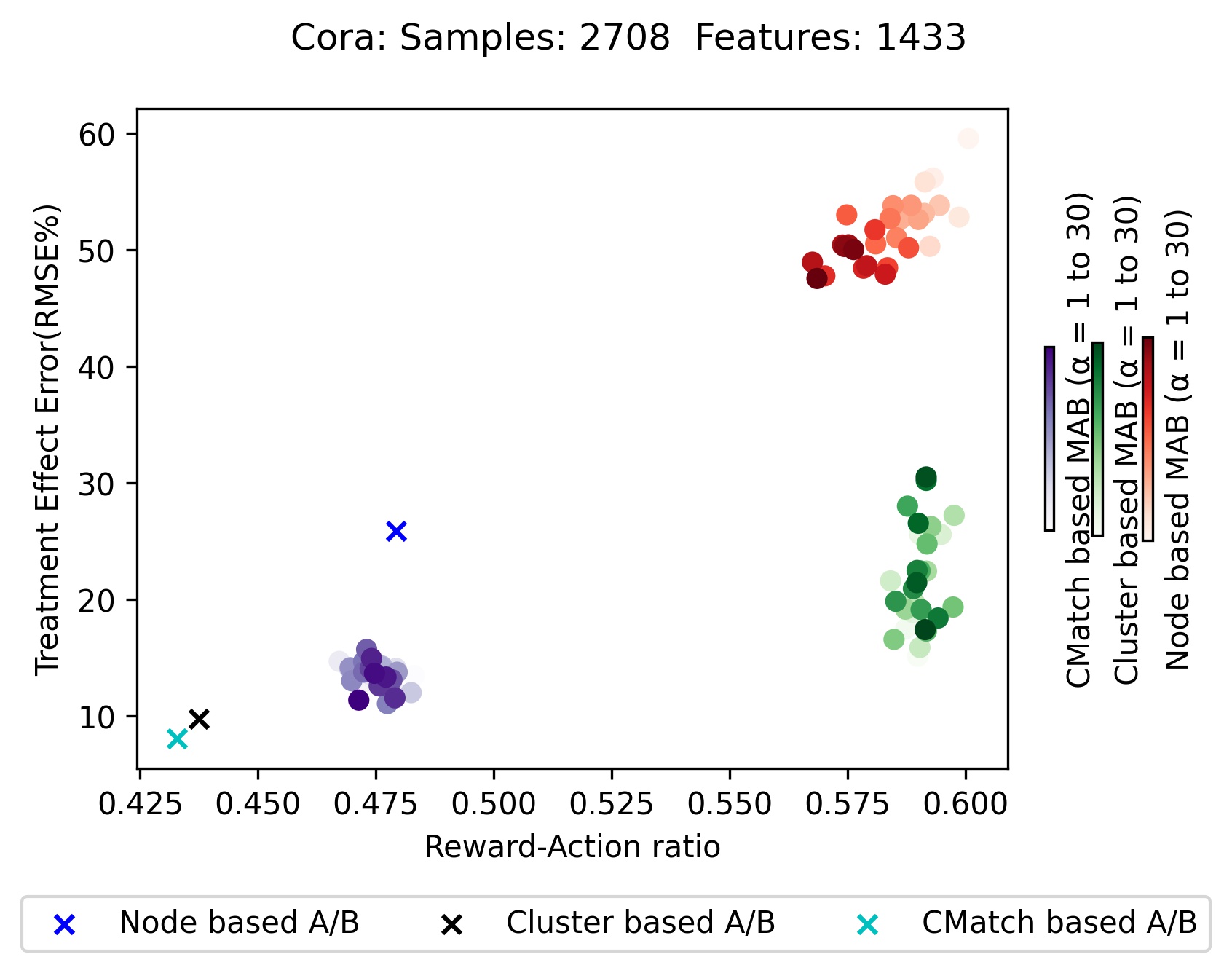}
     \end{subfigure}
     \hfill
     \begin{subfigure}{0.33\textwidth}
         \centering
         \includegraphics[width = 2.2 in]{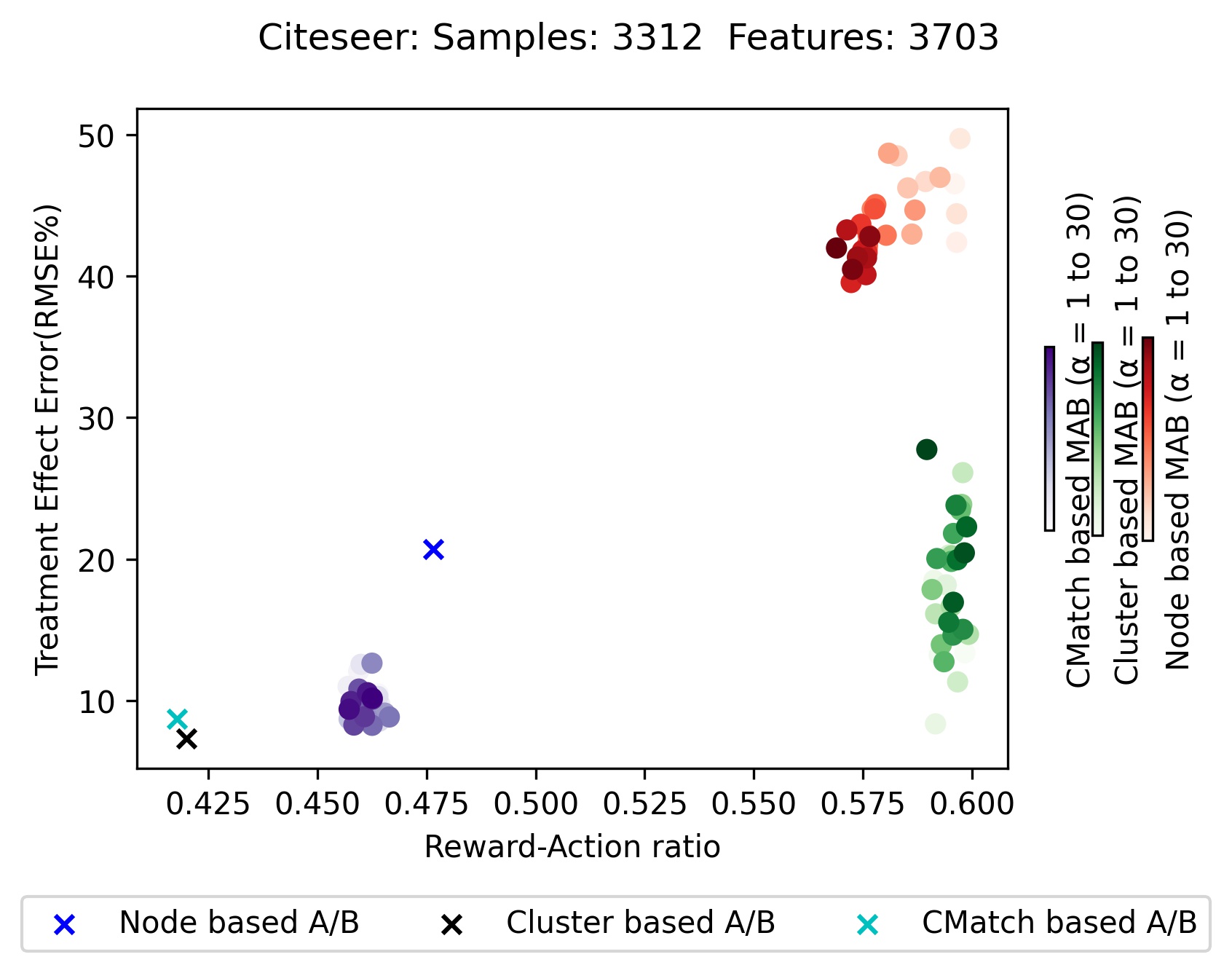}
     \end{subfigure}
     \hfill
     \begin{subfigure}{0.33\textwidth}
         \centering
         \includegraphics[width = 2.2 in]{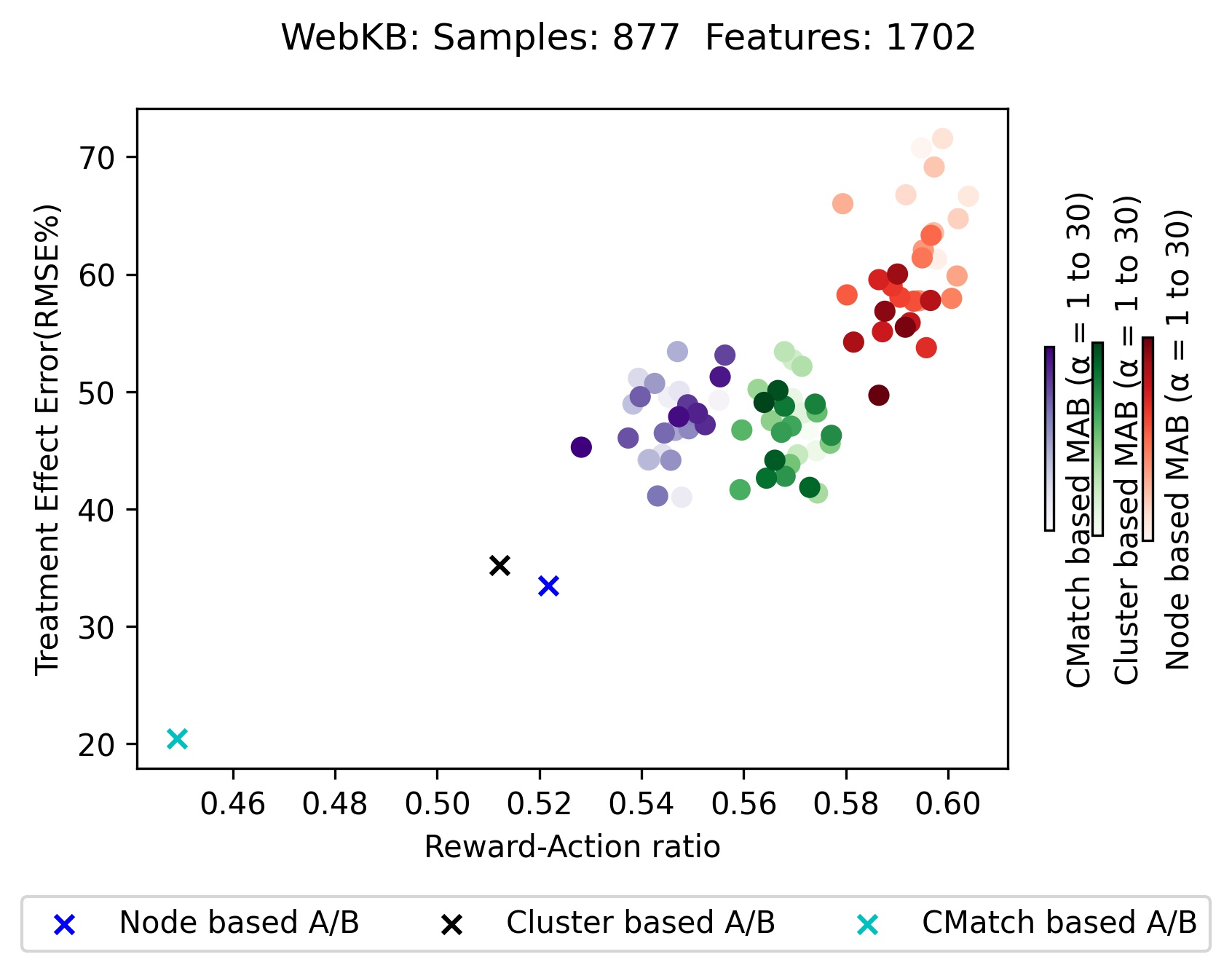}
     \end{subfigure}
     \caption{Plot of reward-action ratio vs RMSE for different values of $\alpha(1$ for low exploration to $30$ for high exploration$)$ where true $TTE = 0.4$}
\end{figure*}
\subsection{CMatch-based MAB}
Here we describe our CMatch-based MAB algorithm as shown in the Algorithm $3$. The graph is clustered first and each cluster is matched and that treatment assignment happens at the matched pair level.

When a new nodes arrives, its cluster is checked to find whether any other node in the same cluster is already assigned to an arm. If any node in that cluster is already assigned to an arm, the newly arrived node is assigned to the same arm [Line: $3-5$]. If no node in that cluster is assigned to an arm, the corresponding matching cluster is checked to see whether any node in that cluster is assigned to an arm. If a node in the matching cluster is assigned to the treatment arm, the newly arrived node is assigned to the control arm, and vice versa [Line: $7-8$]. Otherwise, the newly arrived node is assigned to an arm based on Equation -$3$ [Line: $15-17$]. The learning parameters and metrics of interest are updated for the newly arrived node as shown in the Algorithm $3$ [Line: $21-23$]. 

\section{Experiments}
In this section, we evaluate and show the advantages of our proposed MAB algorithm variants compared to the A/B algorithm variants. To reduce the computational complexity, we use the same cluster list and matching cluster mates/pairs in all of our experiments for the same dataset. 

\subsection{Experimental setup}
For all of our network experiments, we assume the interference between $v_i$ and $v_j$, $e_{v_{i}v_{j}}.p$ is dependent on the their cosine similarly and we consider one-hop interference. We assume that the underlying probability of activating a node due to treatment and allowable interference in the treatment arm/group is $0.6$ and the underlying probability of activating a node due to treatment and allowable interference in the control arm/group is $0.2$ which results in the true causal effect, $TTE = 0.4$. Each node in the explored graph is assigned to be activated or not due to the treatment effect and allowable interference based on these probabilities. We simulate contagion interference considering values based on the edge weights $e.p$. The inactive treated and untreated nodes get activated with the unallowable spillover probability of $e.p$ for each adjacent active node from different treatment class. The nodes arrives at random in all of our experiments.

For graph clustering, we use unweighted Markov Clustering Algorithm (MCL)\cite{Enright02}. 

Cosine similarity has been used to find the similarity between any two nodes. The maximum value for similarity is $1$, which refers to the similarity between two exact same nodes. Node $v_k$ in cluster $c_i$ is matched with node $v_l$ from a different cluster $c_j$ if the pairwise similarity of nodes $sim(v_k, v_l) > \gamma$. We set the threshold based on the covariate distribution of the dataset and consider the second quartile of pairwise similarity for $\gamma$. The average similarity between matched nodes across two clusters $c_i$ and $c_j$ has been considered to find cluster similarity. The cluster $c_i$ is matched with cluster $c_j$ if their weight $w_{i,j}> \beta$. We set the threshold based on the covariate distribution of the dataset and consider the second quartile of pairwise similarity for $\beta$.

 

\subsection{Data}
We consider three real-world attributed network datasets -Cora \cite{mccallum2000}, Citeseer\cite{Sen08}, and WebKB\cite{craven1998}. The Cora dataset is a citation network with binary bag-of-words attributes for each article. This dataset contains $2708$ nodes, $5278$ edges, and $1433$ attributes. Similarly, the CiteSeer dataset is also a citation network which contains $3312$ nodes, $4732$ edges, and $3703$ attributes. Finally, the WebKB is another citation network which contains $877$ nodes, $1608$ edges, and $1702$ attributes. The WebKB network is a combination of four different citation networks, which are disjoint from each another. The unweighted MCL clustering algorithm generates $599$, $1031$, and $152$ clusters for Cora, Citeseer, and WebKB datasets, respectively.
\begin{figure*}[!ht]
     \centering
     \begin{subfigure}{\textwidth}
         \centering
         \includegraphics[width = 3.3 in ]{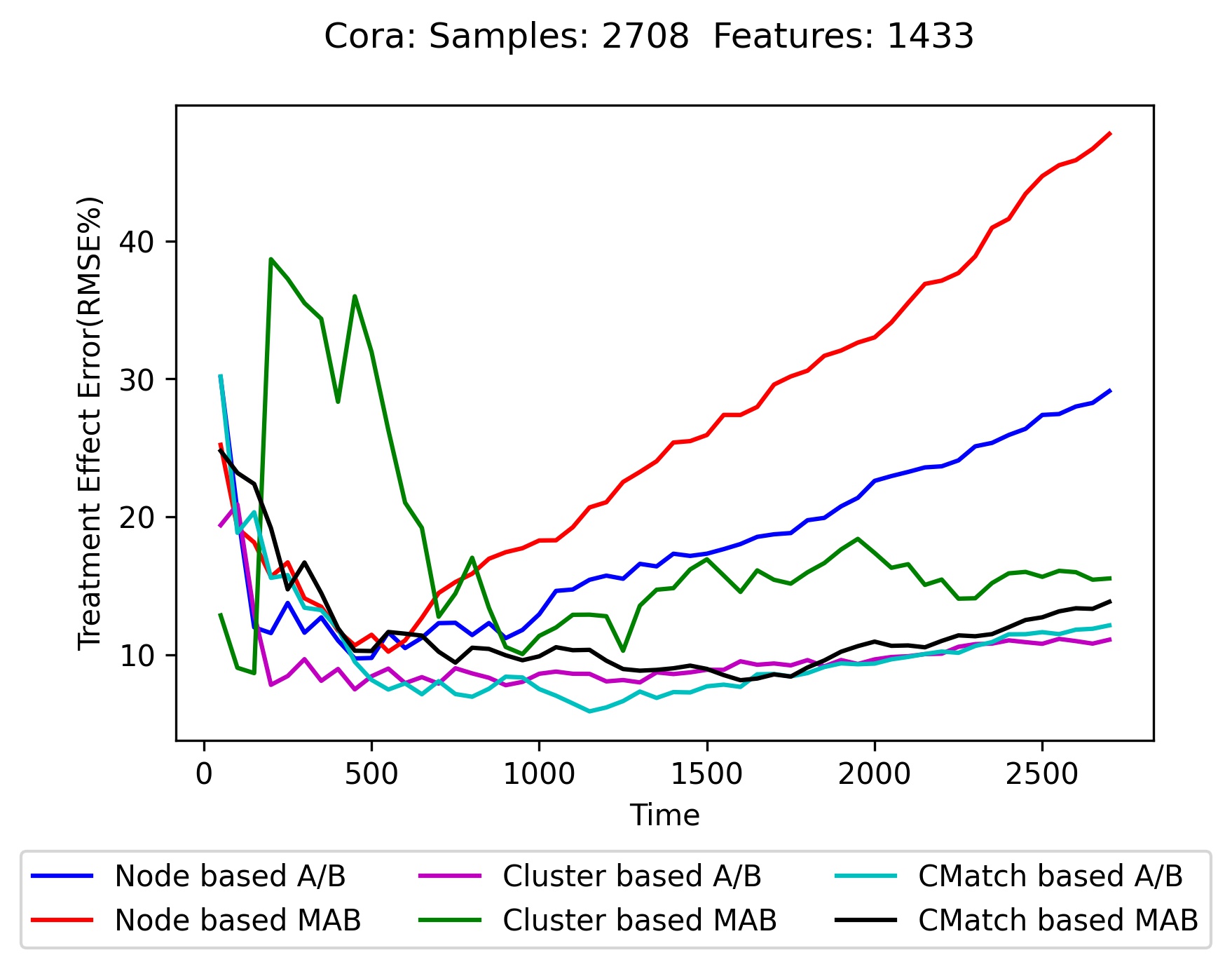}
         \includegraphics[width = 3.3 in]{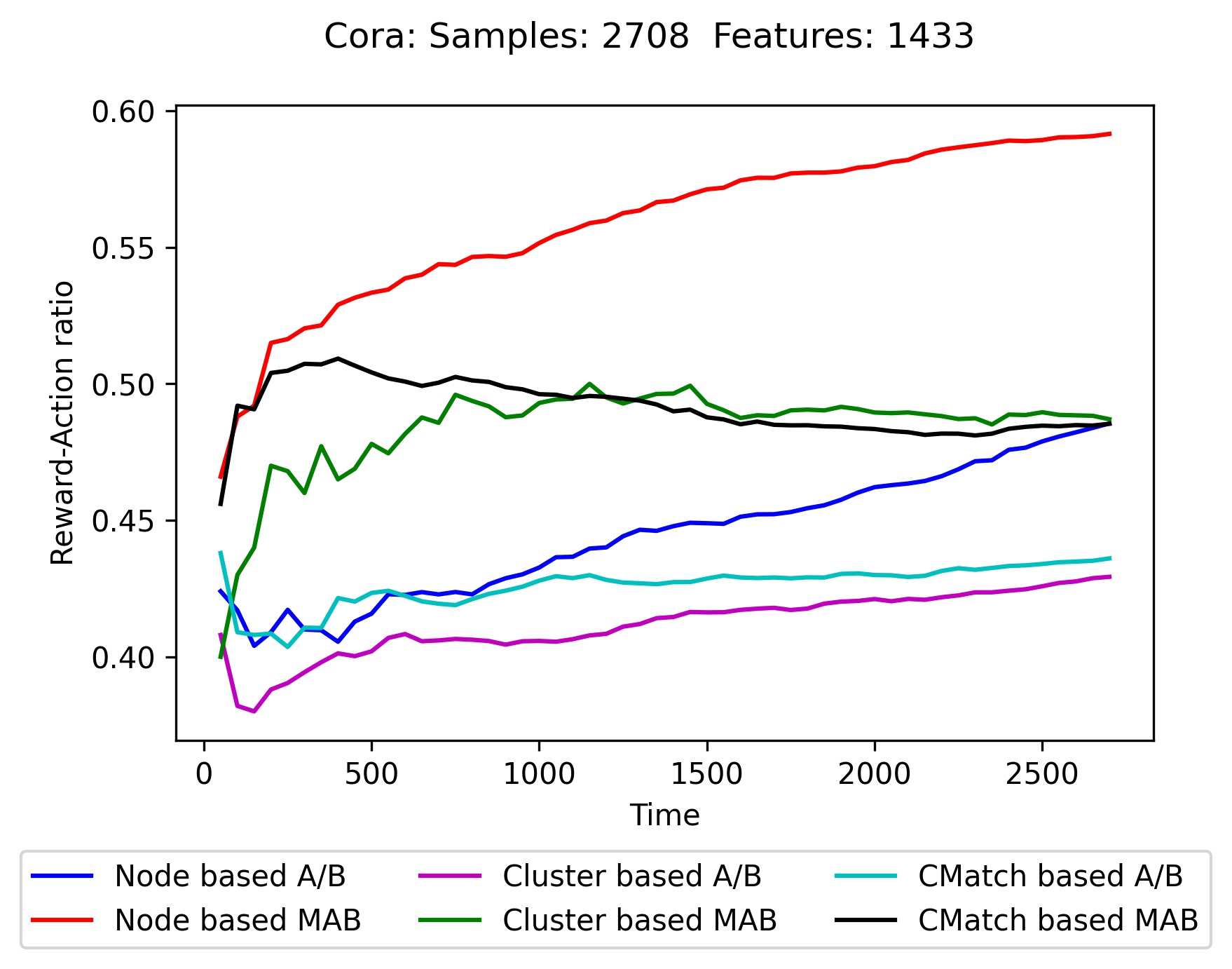}
     \end{subfigure}
     \hfill
     \begin{subfigure}{\textwidth}
         \centering
         \includegraphics[width = 3.3 in]{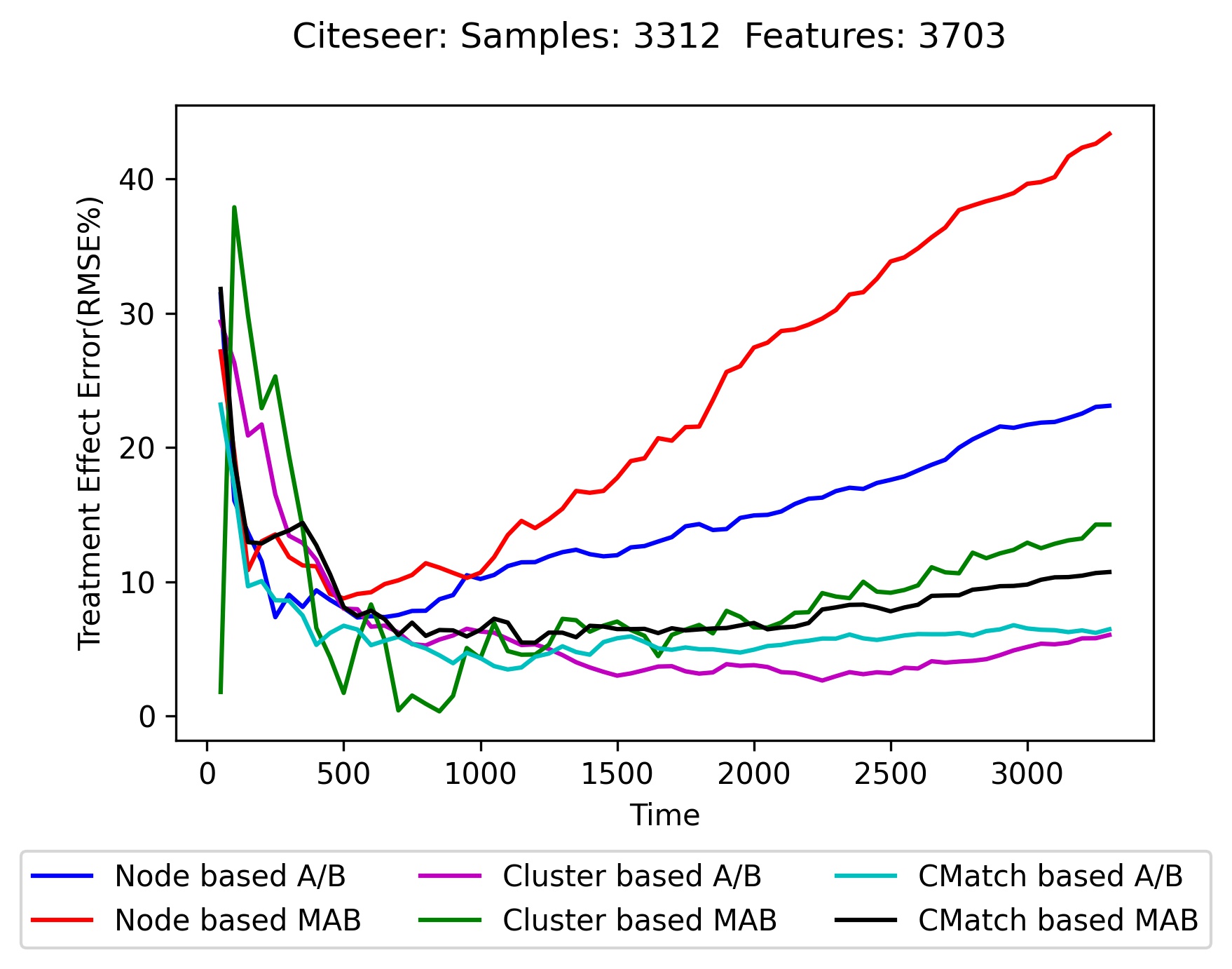}
         \includegraphics[width = 3.3 in]{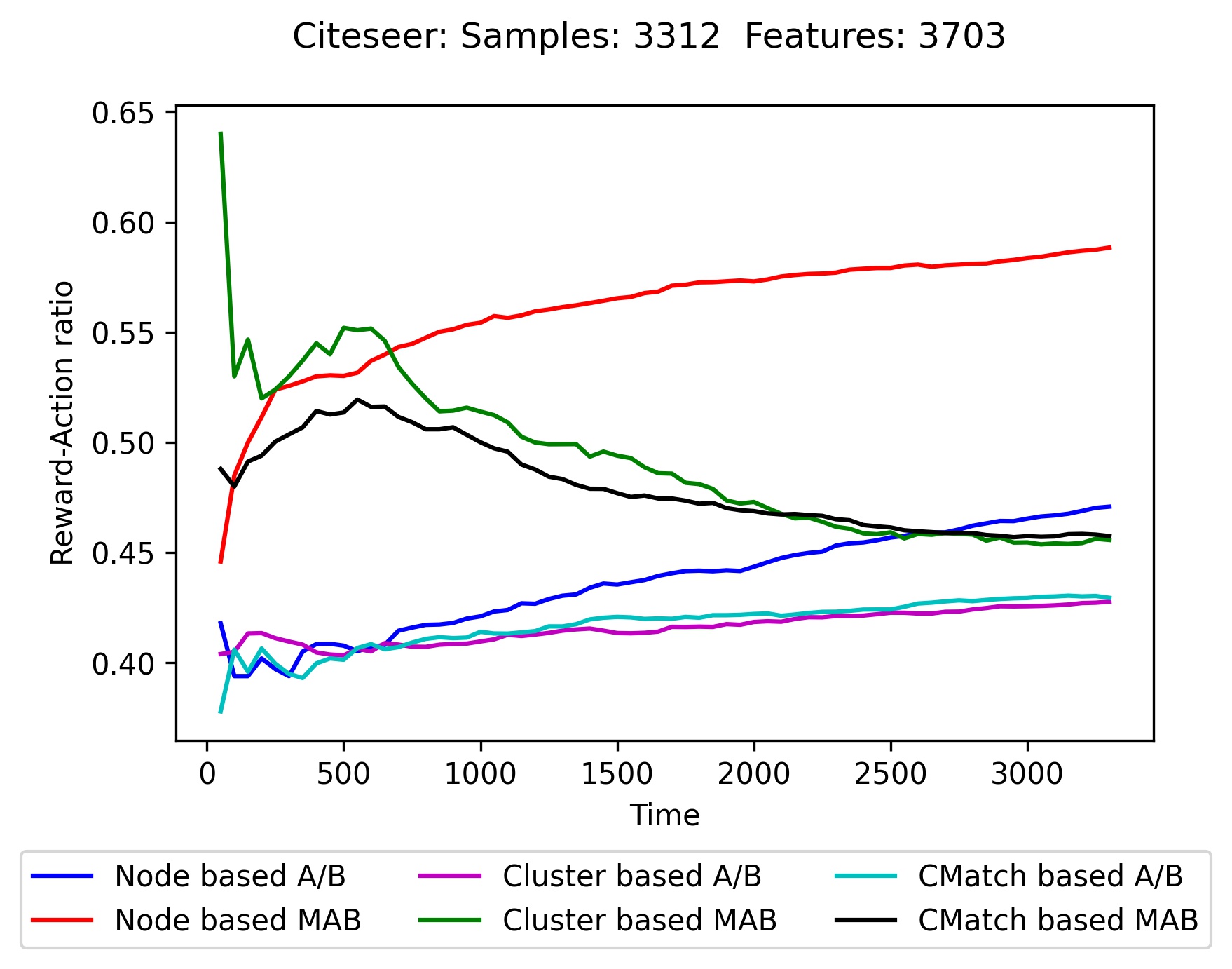}
     \end{subfigure}
     \begin{subfigure}{\textwidth}
         \centering
         \includegraphics[width = 3.3 in]{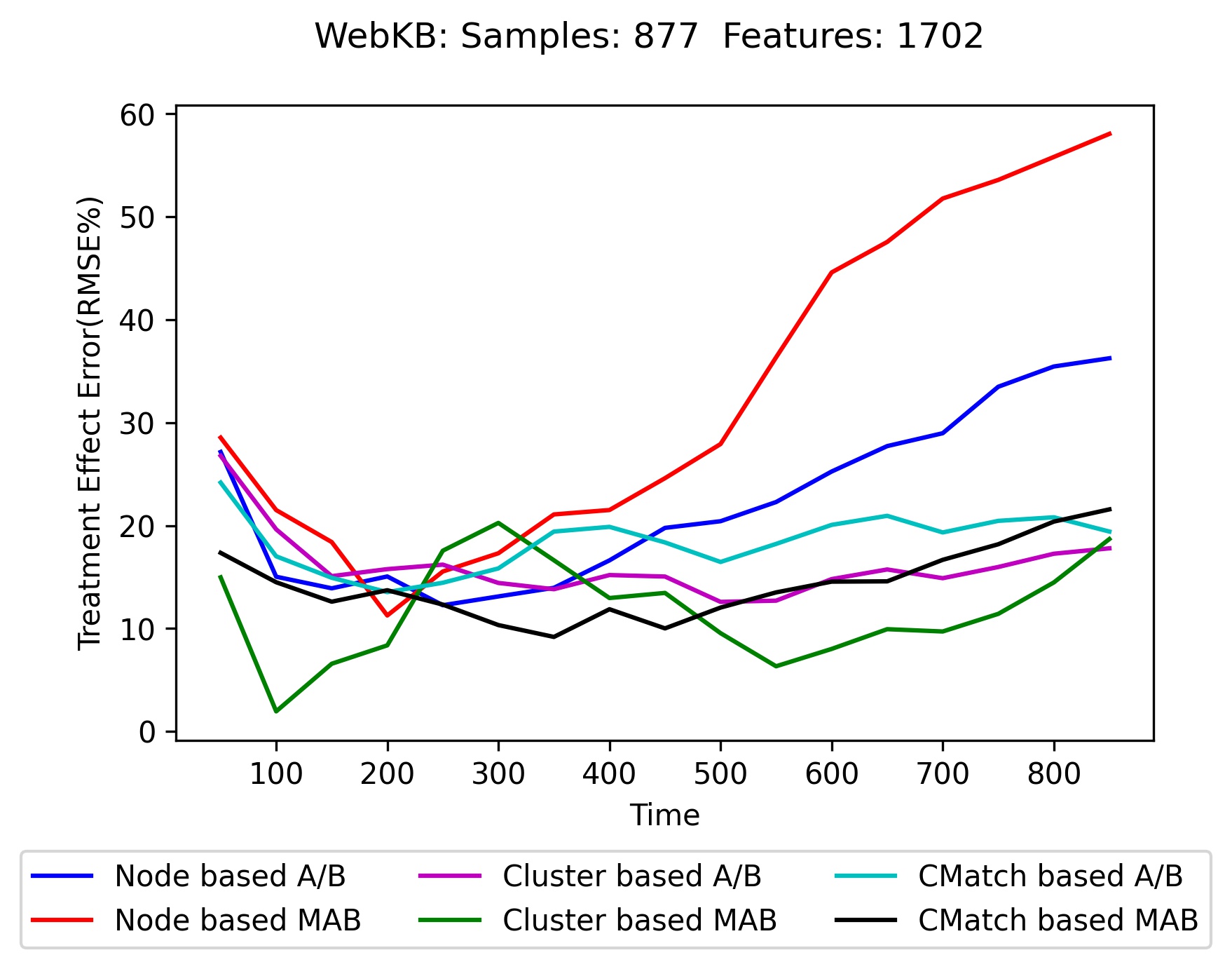}
         \includegraphics[width = 3.3 in]{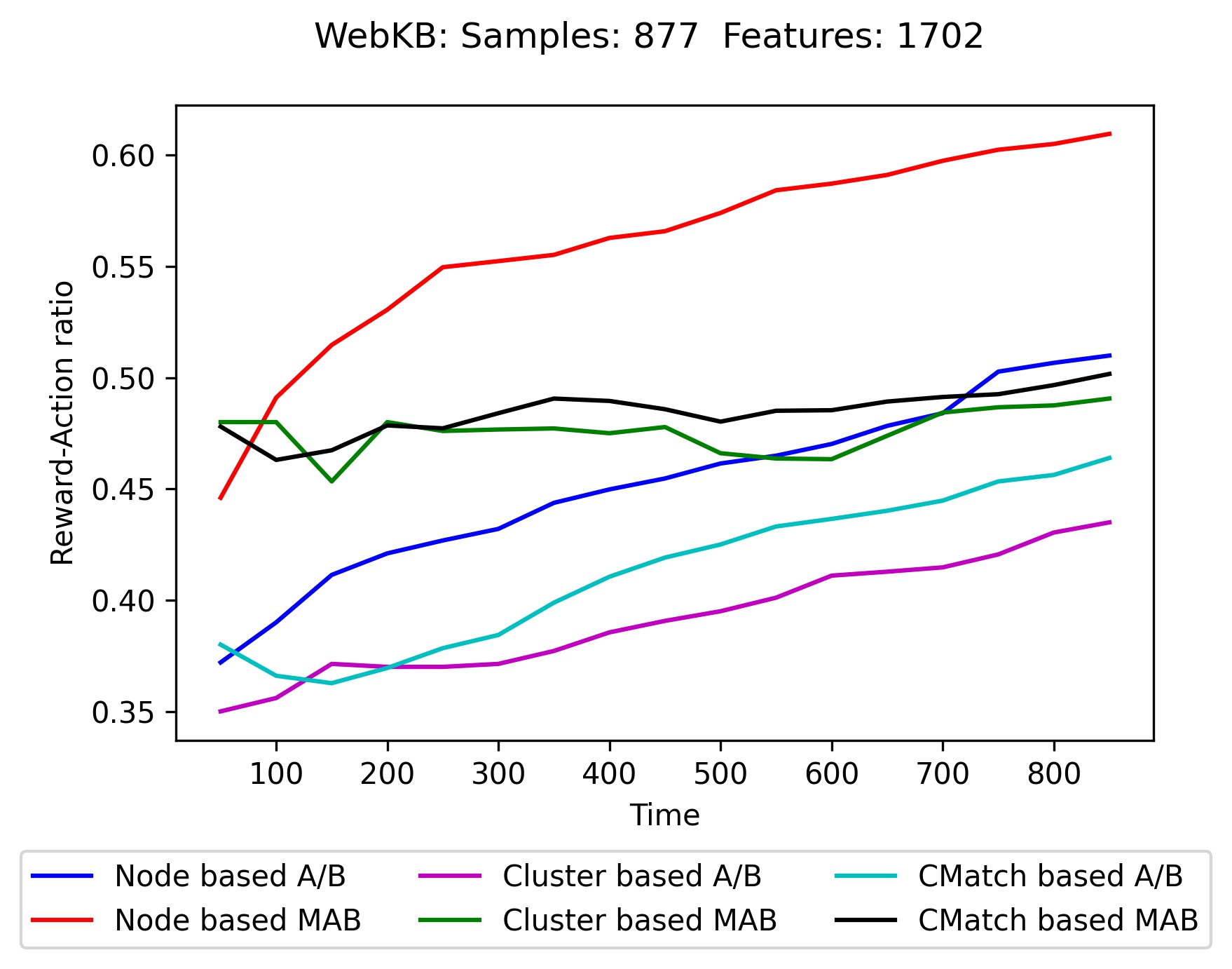}
     \end{subfigure}
     \caption{Plot of metrics for Cora, Citeseer, and WebKB datasets where true $TTE = 0.4$ and $\alpha = 8$}
\end{figure*}
\subsection{Metric}
The metrics of our interest are the root mean squared error (RMSE) and reward-action ratio. 

The RMSE of the total treatment effect after $S$ runs is calculated as follows:
\begin{equation}
    RMSE = \sqrt{\frac{1}{S}\sum_{s=1}^{S} (TTE_s -  \hat{TTE_s})^2}
\end{equation}

The $TTE_s$ and $\hat{TTE_s}$ are the true and estimated TTE in run $s$, respectively. The cumulative RMSE error is reported in terms of the percentage to the true $TTE = 0.4$.

The reward-action ratio is the ratio of total number of activated node to the total number of explored nodes in the network.

We have run all experiments $10$ times and the metrics of interest are averaged over total number of runs.

The cumulative average metrics of interest are reported and plotted for each dataset in Fig. $2$. in a period of $50$ trials or node arrivals for the MAB based variants. To make the plots of the AB based variants compatible to that of MAB based variants, the metrics for the AB based variants are also calculated iteratively and reported in a period of $50$ trials or node arrivals.


\subsection{Results}
\subsubsection{Trade-off between exploration and exploitation:}

In all of our MAB algorithms, $\alpha$ controls the trade-off between exploration and exploitation. The exploration increases with the increase in $\alpha$. Given a range of integer values for $\alpha \in [1, 30]$, we first study the trade-off between TTE error and reward-action ratio in the node-based, cluster-based, and CMatch-based MAB experiments for different datasets. Fig. $1$ shows the TTE error vs reward-action ratio after exploring the whole network with the increase in $\alpha$ from $1$ to $30$. The intensity of color in Fig. $1$ increases with the increase in $\alpha$ in all of our MAB experiments. The change in $\alpha$ has no effect on the A/B experiments.   

These figures clearly show that the reward-action ratio increases at the expense of TTE error for all datasets in the node-based MAB experiments and both the reward-action ratio and TTE error increases with the decrease in $\alpha$. For example, by decreasing the $\alpha$ from $30$ to $1$ in Cora, the reward-action ratio increases from $0.56$ to $0.62$, and TTE error increases from $47\%$ to $59\%$ in the node-based MAB experiments. We need to choose an optimum value of $\alpha$ to make a trade-off between reward-action ratio and TTE error for the node-based MAB experiment.

The cluster-based and CMatch-based MAB experiments do not generate similar pattern of results as node-based MAB experiments. As shown in Fig. $1$, the reward-action ratio does not fluctuate much with the change of $\alpha$ in cluster-based and CMatch-based MAB experiments. For example, the reward-action ratio lies between $0.58$ and $0.6$ for the cluster-based MAB experiment in the Cora dataset. For the CMatch-based MAB experiment, the reward-action ratio lies between $0.46$ and $0.48$ with the change of $\alpha$ from $1$ to $30$ in the same dataset.

The TTE error fluctuates by around $5-10\%$ with the change of $\alpha$ in the CMatch-based MAB experiments for different datasets. For example, the TTE error lies between $9\%$ to $16\%$ in the CMatch-based MAB experiment for the Cora dataset as shown in Fig. $1$. However, in case of cluster-based MAB experiments, TTE error shows a bit more fluctuation with the change in $\alpha$ for all the datasets. 




Based on these results of Fig. $1$, we consider the same $\alpha = 8$ for the node-based, cluster-based and CMatch-based MAB experiments to keep the TTE error minimum and reward-action ratio maximum and compare the performance with the corresponding A/B based experiments. 

\subsubsection{Evaluation of metrics in A/B experiments and MAB experiments for $\alpha = 8$:}

The MAB based algorithms clearly outperform the corresponding A/B based algorithms in terms of reward-action ratio for all the three datasets as shown in the Fig. $2$. For example, in the Citeseer dataset, the node-based MAB shows a reward-action ratio of $0.59$ while node-based A/B shows $0.47$ after exploring the whole network. The cluster-based and CMatch-based MAB experiment show a reward-action ratio of $0.45$ and $0.46$, respectively, while both of their corresponding A/B based experiment show a reward-action ratio of $0.42$ after exploring the whole network as shown in Fig. $2$.

However, the node-based MAB experiment generates much higher error compared to the node-based A/B experiment since more nodes are being assigned to the treatment arm than to the control arm and thus result in more contagion. For example, in the Cora dataset, node-based MAB experiment shows a TTE error of $47.76\%$ while node-based A/B  experiment shows $29.11\%$ after exploring the whole network. However, the cluster-based and CMatch-based MAB methods show almost similar or a bit higher error compared to their corresponding A/B methods. For example, the cluster-based and CMatch-based MAB experiments show a TTE error of $15.52\%$ and $13.83\%$, respectively, while both of their corresponding A/B based experiment show a TTE error of $11.08\%$ and $12.12\%$, respectively, after exploring the whole network as shown in Fig. $2$.

Meanwhile, both the TTE error and reward-action ratio tend to fluctuate less after exploring the first $30-40\%$ of the total network for some datasets (Cora and Citeseer) and thus seem to converge for the cluster-based and CMatch-based MAB experiments as shown in Fig. $2$. However, the same does not go with the other dataset (WebKB).

\section{Discussion}
We demonstrate the trade-off between the TTE error and reward-action ratio, as well as the significance of balance between exploration and exploitation in our MAB algorithms. Our experiments on three network datasets show that node-based MAB algorithms increase reward-action ratio at the expense of significant increase in TTE error, but cluster-based and CMatch-based MAB algorithms are shown to increase reward-action ratio significantly when compared to their A/B testing counterparts with very small increase in TTE error. Moreover, both metrics tend to become stable after exploring $30-40\%$ of the networks and seem to converge. Some possible extension of our work is to incorporate heterogeneous node outcomes based on their features and evaluate the performance on a large variety of network datasets.



\bibliography{aaai22}

\end{document}